\documentclass[lettersize,journal]{IEEEtran}
\usepackage{amsmath,amsfonts}
\usepackage{algorithmic}
\usepackage{array}
\usepackage[caption=false,font=normalsize,labelfont=sf,textfont=sf]{subfig}
\usepackage{textcomp}
\usepackage{stfloats}
\usepackage{url}
\usepackage{verbatim}
\usepackage{graphicx}
\def\BibTeX{{\rm B\kern-.05em{\sc i\kern-.025em b}\kern-.08em
    T\kern-.1667em\lower.7ex\hbox{E}\kern-.125emX}}
\usepackage{balance}
\usepackage{pifont}
\usepackage[normalem]{ulem}  
\usepackage{algorithm}

\usepackage{wrapfig}
\usepackage{booktabs} 
\usepackage{tabularx}
\usepackage{xcolor}
\usepackage{threeparttable} 

\usepackage{hyperref}
\usepackage{cleveref}

\newboolean{debug}
\setboolean{debug}{false} 
\usepackage{tabularx, booktabs, pifont, threeparttable}
\usepackage{cancel}

\newcommand{\msout}[1]{%
    \ifthenelse{\boolean{debug}}{%
        \sout{#1}%
    }{%
    }%
}

\newcommand{\lia}[1]{%
    \ifthenelse{\boolean{debug}}{%
        \textcolor{blue}{#1}
    }{%
        #1%
    }%
}

\newcommand{\zou}[1]{%
    \ifthenelse{\boolean{debug}}{%
        \textcolor{red}{#1}
    }{%
        #1%
    }%
}

\newcommand{\reffig}[1]{Figure \ref{#1}}

\newcommand{\refeq}[1]{Equation (\ref{#1})}
\newcommand{\refeqn}[1]{(\ref{#1})}
\newcommand{\refsec}[1]{Section \ref{#1}}

\setboolean{debug}{true} 
\setboolean{debug}{false} 

\begin{document}
\title{VisFly-Lab: Unified Differentiable Framework for First-Order Reinforcement Learning of Quadrotor Control}
\author{Anonymous}
\author{Fanxing Li, Fangyu Sun, Tianbao Zhang,  Shuyu Wu, Dexin Zuo, yufei Yan, Wenxian Yu, Danping Zou*}



\maketitle


\begin{abstract}
First-order reinforcement learning with differentiable simulation is promising for quadrotor control, but practical progress remains fragmented across task-specific settings. To support more systematic development and evaluation, we present a unified differentiable framework for multi-task quadrotor control. The framework is wrapped, extensible, and equipped with deployment-oriented dynamics, providing a common interface across four representative tasks: hovering, tracking, landing, and racing. We also present the suite of first-order learning algorithms, where we identify two practical bottlenecks of standard first-order training: limited state coverage caused by horizon initialization and gradient bias caused by partially non-differentiable rewards. To address these issues, we propose Amended Backpropagation Through Time (ABPT), which combines differentiable rollout optimization, a value-based auxiliary objective, and visited-state initialization to improve training robustness. Experimental results show that ABPT yields the clearest gains in tasks with partially non-differentiable rewards, while remaining competitive in fully differentiable settings. We further provide proof-of-concept real-world deployments showing initial transferability of policies learned in the proposed framework beyond simulation.

\end{abstract}



\section{Introduction}
Learning-based quadrotor control is attractive because it transfers computation into a fast policy and avoids expensive online optimization during deployment. Among these approaches, first-order reinforcement learning (RL) with differentiable simulation is particularly appealing for quadrotors \cite{n_wiedemann_training_2023}, whose dynamics are smooth, structured, and analytically tractable. This makes quadrotors a natural platform for exploiting analytical gradients through simulation rather than relying only on sampled policy-gradient estimates.

However, practical progress on first-order learning for quadrotor control remains fragmented. Existing differentiable-quadrotor studies \cite{lu_yopov2-tracker_2025,zhang2024back,heeg2025learning} are still largely task-specific, making it difficult to compare optimization behavior across tasks under a shared interface and dynamics model. In addition, several existing studies \cite{zhang2024back,lu_yopov2-tracker_2025} rely on simplified assumptions, introducing extra mismatch between training and deployment. Specifically, it use point-mass kinematics to replace thorough dynamics, unable to be transferred to other configuration. Furthermore, the application of first-order approaches has not been formalized into a concrete algorithm or a practical implementation.
These limitations motivate the need for a unified and deployment-oriented framework for first-order RL of quadrotor control. 

Within such a unified setting, we observe two recurring practical difficulties. First, horizon-based rollout training may under-cover feasible initial states, since many implementations initialize each new horizon from the terminal state of the previous one. Second, when task rewards are partially non-differentiable, backpropagation-through-time (BPTT) \cite{freeman_brax_2021} and its variant \cite{xu_accelerated_2022} can only propagate gradients through the differentiable reward terms, introducing a gradient bias with respect to the full task objective. This issue is particularly relevant in tasks where binary or conditional success-related rewards play an behavioral role. 

To study and address these issues, we develop a unified differentiable framework for multi-task quadrotor control. The framework is wrapped, extensible, and equipped with deployment-oriented differentiable dynamics, providing a common interface across four representative tasks: hovering, tracking, landing, and racing. On top of this framework, we propose Amended Backpropagation Through Time (ABPT), which combines short-horizon differentiable optimization, a value-based auxiliary objective, and visited-state initialization to mitigate the above bottlenecks.

We evaluate the proposed framework and ABPT on the four tasks and compare ABPT against representative baselines. Results show that ABPT is competitive across the unified benchmark and yields the clearest gains in tasks with partially non-differentiable rewards. We also provide proof-of-concept real-world deployments of policies learned in the proposed framework, offering initial evidence of transferability beyond simulation rather than a comprehensive quantitative hardware benchmark.
The contributions are summarized as follows:
\begin{itemize}
    \item We develop a unified, wrapped, and extensible differentiable framework for multi-task quadrotor control, with deployment-oriented dynamics and a shared interface across hovering, tracking, landing, and racing.
    \item We propose the first-order learning suite including BPTT and SHAC wrapped as stable-baselines. Besides, to address the bottlenecks of current pipeline, we present ABPT, which combines differentiable rollout optimization, a value-based auxiliary objective, and visited-state initialization to improve training robustness.
    \item We demonstrate through simulation that ABPT is competitive across the unified benchmark,
    and we provide proof-of-concept real-world deployments showing initial transferability beyond simulation.
\end{itemize}

\section{Related Work}
\begin{figure*}[h]
    \centering
    \includegraphics[width=1.0\textwidth]{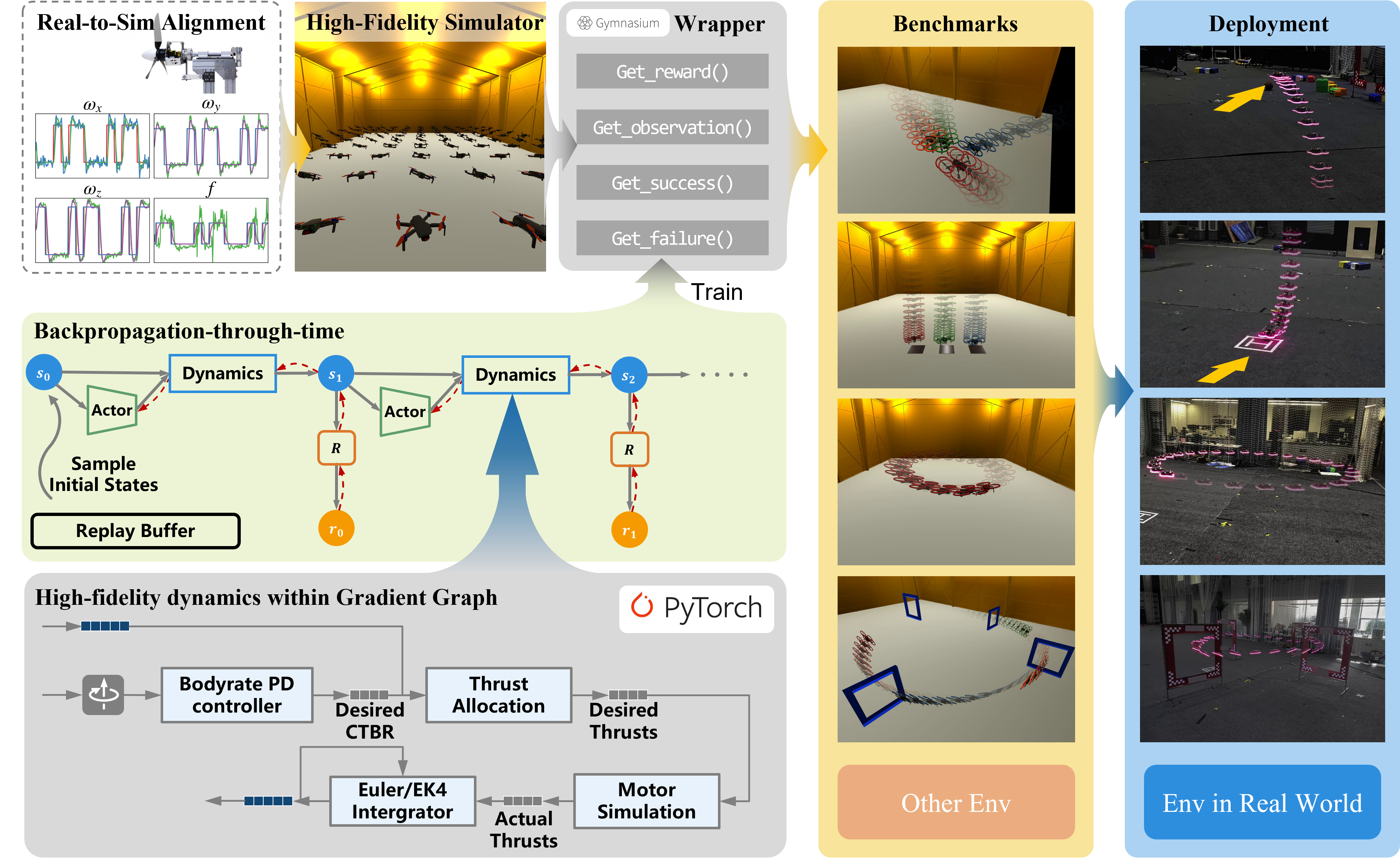}
    \caption{Overview of the proposed unified differentiable framework for multi-task quadrotor control. The framework provides a wrapped and extensible architecture, supports differentiable simulation with deployment-oriented dynamics, and includes four representative tasks: hovering, tracking, landing, and racing.
    }
    \label{fig:overview}
\end{figure*}

\subsection{Differentiable Simulators}
Policy learning via differentiable physics is an approach that integrates the physical simulations with differentiable dynamics to enable policy learning directly by using gradient-based optimization. Making the dynamics differentiable in the simulator is the key to this approach. 
DiffTaichi \cite{hu_difftaichi_2020} is a comprehensive differentiable physics engine that includes simulations of fluid, gas, rigid body movement, and more. 
In the field of robotics, Brax \cite{freeman_brax_2021} offers differentiable versions of common RL benchmarks, built on four physics engines, including JAX and MuJoCo \cite{todorov_mujoco_2012}. Another line of research focuses on addressing challenges in contact-rich environments. For example, Heiden et al. \cite{heiden_neuralsim_2021} tackle the contact-rich discontinuity problem in quadruped robots by employing a neural network to approximate the residuals. Dojo \cite{howell_dojo_2023} enhances contact solvers and integrates various integrators to accelerate computations while maintaining fidelity. 
VisFly \cite{li_visfly_2024} introduces a versatile drone simulator with fast rendering, based on Habitat-Sim \cite{savva2019habitat}, providing a platform for high-level applications. To enhance the efficiency, many simulators leverage GPU-accelerated frameworks like JAX \cite{schoenholz_jax_2020} and PyTorch \cite{paszke2017automatic} for faster computations.

\subsection{First-order Gradient Training}
With the differentiable simulators, the policy can be trained through BPTT by using the first-order gradients. Though first-order gradients enable faster and more accurate gradient computation, they suffer from gradient explosion/vanishing or instability caused by smooth dynamics. Many attempts have tried to address these issues and strengthen robustness. PODS \cite{mora_pods_2021} leverages both first- and second-order gradients with respect to cumulative rewards. SHAC \cite{xu_accelerated_2022} employs an actor-critic framework, truncates the learning window to avoid vanishing/exploding gradients, and smooths the gradient updates.
AHAC \cite{georgiev2024adaptivehorizonactorcriticpolicy} makes the horizon adaptive to reduce sampling error in scenarios involving stiff dynamics.
SAPO \cite{xing2024stabilizing} introduces entropy to strengthen the training stability especially in soft-body simulation. Generally, up to now SHAC and BPTT are the most widely recognized methods for training with first-order gradients.

\subsection{Studies on Quadrotors using Analytical Gradient}
Eliminating physical contact with external objects, the dynamics of quadrotors can be the smoothest among all robotic systems. As a result, it becomes the most suitable platform for training with first-order gradients.  \cite{n_wiedemann_training_2023} compared BPTT, PPO, and MPC for trajectory tracking of quadrotors, showing that BPTT can achieve same performance as MPC but with much less inference time. \cite{heeg2025learning} trained a point-feature based policy for stable hovering. \cite{pan_learning_2025} leverages the fast convergence property of BPTT to enhance the adaptation of quadrotors under different winds. \cite{zhang2024back} firstly introduces image into observations, and achieves high-speed collision-free flight. \cite{lu_yopov2-tracker_2025} also trains a policy for obstacle avoidance, but this policy generates trajectories instead of direct control commands. 
All these works treat analytical gradients primarily as a tool for solving specific tasks, rather than organizing the training process into a unified framework for downstream development. In addition, several of them rely on kinematic models instead of deployment-oriented dynamics, which introduces an additional mismatch between training and deployment beyond sim-to-real gap.

\section{Unified Differentiable Framework for Multi-Task Quadrotor Control}

\subsection{Preliminaries}
%
The goal of reinforcement learning is to find a stochastic policy $\pi$ that maximizes the expected cumulative reward, or the expected return, over a trajectory $\tau$. 
\zou{In a common actor-critic pipeline, both the actor $\pi_\theta$ and the critic -- either the value function $V_{\phi}(s)$
or the state-value function $ Q_{\phi}(s, a)$ -- are approximated by neural networks with parameters $\theta$ and  $\phi$. The key problem is how to estimate the gradients to optimize the expected return. The methods could be divided into two following categories: }

\textbf{Policy Gradient.}
Policy gradient methods estimate the gradient of the expected return using the log-probability of sample trajectories, conditioned on the policy's action distribution. Given a batch of experience, the policy gradient is computed as:
\begin{equation}
\label{eq:policy_grad}
    \nabla^{[0]}_\theta \mathcal{J}_\theta = \frac{1}{|\mathcal{B}|} \left[ \sum_{\tau\in \mathcal{B}}\sum_{t=0}^T \nabla_\theta \log \pi_\theta(a_t \mid s_t) A^{\pi_\theta}(s_t, a_t) \right],
\end{equation}
where $A^{\pi_\theta}(\cdot)$ represents the advantage derived from the value functions using current policy, $\mathcal{B}$ denotes the minibatch of sampled trajectories, $\tau$ represents a trajectory within the minibatch. 
Because this formulation does not require differentiating through the environment dynamics, it is also named {zeroth-order gradient (ZOG)}.

\textbf{Value Gradient.}
Value gradient methods compute the policy gradient by differentiating through the action-value function:
\begin{equation}
\label{eq:value_grad}
    \nabla^{[q]}_\theta \mathcal{J}_\theta = \frac{1}{|\mathcal{B}|} \left[ \sum_{i=1}^{|\mathcal{B}|} \nabla_\theta Q_{\phi}\big(s^{i}, \pi_\theta(s^{i})\big) \right]
\end{equation}
\cite{gao_adaptive_gradient_2024} named this gradient estimator as {Q gradient (QG)}. Compared with {ZOG}, the accuracy of value-function approximation is particularly critical for actor training, since {QG} relies directly on backpropagation through the action-value function. In contrast, {ZOG} estimates advantages with respect to the current policy, which makes actor training more robust to imperfections in critic learning.

\label{sec:no-diff-rewards}
\textbf{First-order Gradient.} Given the state dynamics $T$ and reward function $R$ being differentiable, one can compute the exact gradients of the expected return for policy learning via backpropagation through time. This exact gradient estimate is called {first-order gradient (FOG)}:
\begin{equation}
\label{eq:BPTT}
\nabla_\theta \mathcal{J}_\theta = \left( \sum_{k=1}^{N} \gamma^k \frac{\partial R(s_{k},a_{k})}{\partial \theta} \right),
\end{equation}
where $N$ represents the horizon length, $k$ denotes the $k$-th step within the trajectory, and $R$ represents the reward function.
Considering computation relationship between policy weights and reward function, \refeq{eq:BPTT} could be expanded as:
\begin{equation}
\begin{aligned}
        \nabla_\theta J_\theta
&=\sum_{k=1}^{N}\gamma^k \Bigg(\frac{\partial R(s_k,a_k)}{\partial a_k}\frac{\partial a_k}{\partial \theta} +\\ 
&\frac{\partial R(s_k,a_k)}{\partial s_k}\sum_{i=1}^k \Big [ \Big( \prod_{j=i+1}^k \frac{\partial s_j}{\partial s_{j-1}} \Big) \frac{\partial s_i}{\partial a_i}\frac{\partial a_i}{\partial \theta} \Big ] \Bigg)
\end{aligned}
\end{equation}
where $\partial s_j/\partial s_{j-1}$ is easily inferred by First-Order Ordinary Differential Equation and transition function:
\begin{equation}
\begin{aligned}
\dot{s}&=F(s, a)     \\
    s_j &= F(s_{j-1},a_{j-1})dt + s_{j-1}
\end{aligned}
\end{equation}
We will introduce the differential equations $F$ in next subsection.
\subsection{High-Fidelity Differentiable Simulation}
\label{sec:dynamics}
We deliberately adopt PyTorch for our implementation, as opposed to more widely used differentiable implementations  such as CUDA or JAX in the context of MuJoCo. Although its computational speed is marginally slower than the alternatives, PyTorch offers a more accessible and user-friendly programming interface, owing to its widespread adoption and familiarity within the academic community.

Quadrotor dynamics aligned with real-world conditions are considerably more complex than those typically assumed in simulation. The dynamics are modeled in full 6-DoF to capture the complex interactions between translational motion, rotational dynamics, aerodynamic drag, and actuator dynamics. Specifically, the state evolution is governed by:
\begin{equation}
\small
\begin{aligned}
    &\dot{\mathbf{x}}_W = {\mathbf{v}}_W, \quad 
    \dot{\mathbf{v}}_W = \tfrac{1}{m} \mathbf{R}_{WB}(\mathbf{f} + \mathbf{d}) + \mathbf{g}, \\
    &\dot{\mathbf{q}} = \tfrac{1}{2} \mathbf{q} \otimes \mathbf{\Omega}, \quad
    \dot{\mathbf{\Omega}} = \mathbf{J}^{-1} (\boldsymbol{\tau} - \mathbf{\Omega} \times \mathbf{J} \mathbf{\Omega}),
\end{aligned}
\end{equation}
where the translational states $(\mathbf{x}_W, \mathbf{v}_W)$, orientation quaternion $\mathbf{q}$, and angular velocity $\mathbf{\Omega}$ evolve under the influence of gravity $g$, collective thrust vector $\mathbf{f}$, torque $\boldsymbol{\tau}$,and drag force $\mathbf{d} $. The quaternion product is denoted by $\otimes$, and $\mathbf{R}_{WB}$ is the rotation matrix from body to world frame. $m$ and $\mathbf{J}$ respectively denote mass and inertial matrix.

The aerodynamic drag $\mathbf{d}$ is modeled as quadratic in body-frame velocity:
\begin{equation}
\small
    \mathbf{d} = \tfrac{1}{2}\rho \, \mathbf{v}_B \odot \mathbf{v}_B \, \mathbf{C}_d \odot \mathbf{s},
\end{equation}
where $\rho$ is the air density, $\mathbf{C}_d$ the drag coefficients, $\mathbf{s}$ the effective cross-sectional areas, and $\mathbf{v}_B$ the velocity in the body frame. The operator $\odot$ denotes element-wise multiplication.

Under collective thrust and bodyrates (CTBR) control, the action $\mathbf{a}$ consists of the collective thrust along z-axis ${f}$ and the desired bodyrates $\Omega^{des}$. The PD controller compute the executed target torque:
\begin{equation}
\small
\boldsymbol{\tau} 
= K_p^\Omega\,(\boldsymbol{\Omega}^{des} - \boldsymbol{\Omega})
\;+\; K_d^\Omega\,(\dot{\boldsymbol{\Omega}}^{des} - \dot{\boldsymbol{\Omega}}),
\end{equation}

Such commands are distributed onto the four individual motors through a control allocation process:
\begin{equation}
\small
\begin{bmatrix}
f_1 \\ f_2 \\ f_3 \\ f_4
\end{bmatrix}
=
\begin{bmatrix}
1 & 1 & 1 & 1 \\
0 & l & 0 & -l \\
-l & 0 & l & 0 \\
c_\tau & -c_\tau & c_\tau & -c_\tau
\end{bmatrix} ^{-1}
\begin{bmatrix}
f \\ \tau_x \\ \tau_y \\ \tau_z
\end{bmatrix},
\end{equation}
where  $\tau_{x,y,z},f_{1,2,3,4}$ are the components of torque $\boldsymbol{\tau}$ and thrust $f$. The matrix $\mathbf{M}$ denotes the allocation matrix that maps individual rotor thrusts to total thrust and body torques. $l$ is the arm length and $c_\tau$ is the rotor torque coefficient.  
This formulation ensures that the collective thrust and commanded bodyrates are consistently mapped to the individual motor thrusts, enabling directly training low-level policies on real quadrotors. 


To account for actuator dynamics, a first-order exponential model with time constant $c$ is introduced to describe the delay between commanded and actual rotor speeds:
\begin{equation}
\small
\omega_i = \omega_i^{des} + (\omega_{i-1}-\omega_i^{des})e^{-c\cdot dt},\quad {f}_i= k_2 \omega_i^2 + k_1 \omega_i + k_0 
\end{equation}
where $\omega_i$ is the rotor speed, $\omega_{i}'$ and $\omega_i^{des}$ are the previous and desired speeds, and $k_2, k_1, k_0$ are thrust coefficients. ${f}_i$ denotes the thrust along the $z$-axis of rotor $i$.
The device communication delay is modeled with a one-step delay:
\begin{equation}
\small
    \mathbf{a}_t = \mathbf{a}_{t-1},
\end{equation}

Then, to reduce simulation-to-reality gap, we made system identification to finetune the parameters in simulation, aligning the control response as similar as possible. Such complexity makes first-order gradient computation in backpropagation particularly challenging.

\begin{table*}[t]
\scriptsize
\centering
\caption{Task Definitions, Observations, and Reward Structures in the Unified Multi-Task Framework}
\label{tab:taskDefine}
\begin{tabularx}{\textwidth}{l  | X | l }
\toprule
\textbf{Environments} & \textbf{Observation} & \textbf{Reward Function}  \\
\midrule
\textbf{Hovering}  & state \& $\hat{p}$ & $c - k_1 \left \|p-\hat{p}\right\| - k_2 \left\|q-\hat{q}\right\| - k_3 \left\|v\right\| - k_4 \left\|\omega\right\|$ (fully DIFF) \\
\textbf{Tracking} & state \& next 10 $\hat{p}_{i=1\sim10}$  & $c- k_1 \left \|p-\hat{p}_0\right\| - k_2 \left\|q-\hat{q}\right\| - k_3 \left\|v\right\| - k_4 \left\|\omega\right\|$ (fully DIFF)\\
\textbf{Landing} & state \& $\hat{p}$ & $-k_1 f^+ \big( \left \|p_{xy}-\hat{p}_{xy}\right\| \big) + k_2 f^+ \big(  \left\|v_z-\hat{v}_z\right\| \big) + k_3 s$ (partially DIFF)\\
\textbf{Racing} & state \& next 2 $\hat{p}_{i=1,2}$ of gates & $c - k_1 \left \|p-\hat{p}_0\right\| - k_2 \left\|q-\hat{q}\right\| - k_3 \left\|v\right\| - k_4 \left\|\omega\right\| + k_5s$ (partially DIFF) \\
\bottomrule
\end{tabularx}

\begin{tablenotes}
\item[1] $c$ represents a small constant used to ensure the agent remains alive. $k_i$ denotes constant weights for different reward contributions, with these weights being distinctly defined for each task. $s$ is a boolean variable that indicates whether the task is successfully completed, to award once at termination if it succeeds. The state comprises position ($p$), orientation ($q$), linear velocity ($v$), and angular velocity ($\omega$). $f^+(\cdot)$ denotes an increasing mapping function used to normalize the reward and $\hat{(\cdot)}$  denotes target status. DIFF is abbreviation for differentiable. All the action types are CTBR.

\end{tablenotes}

\end{table*}

\subsection{Representative Task Set}

\textbf{Hovering.} Starting from a random position, the quadrotor needs to hover stably at a target location. Fully differentiable rewards are used in this task.

\textbf{Tracking.} Starting from a random position, the quadrotor tracks a circular trajectory with a fixed linear velocity. Fully differentiable rewards are used in this task.
Specifically, the quadrotor needs to solve the next 10 points along the trajectory based on its current position.

\textbf{Landing.} Starting from a random position, the quadrotor gradually descends, and eventually lands at the required position on the ground. 
The drone will crash if it lands with a speed beyond the safe threshold, making it challenging to find the balance during exploration. 

\textbf{Racing.} The quadrotor flies through four static gates as quickly as possible in a given order repeatedly. 
The drone starts at a random position among the gates and flies through them repeatedly in a specified order. 


Table I summarizes the observations and reward structures of the four tasks, which span different levels of complexity and differentiability.
In contrast, both the landing and racing tasks incorporate binary rewards.
However, there is a key difference between them. In landing, the continuous reward teaches the quadrotor to gradually slow down and descend, while the binary reward serves only to confirm successful touchdown. In racing, however, the binary reward plays a decisive role by preventing the quadrotor from hovering near the gates without actually passing through them.




\section{Amended Backpropagation-through-time}

\subsection{Practical Bottlenecks of First-Order RL}
\begin{figure*}[h]
    \centering
    \includegraphics[width=1.0\textwidth]{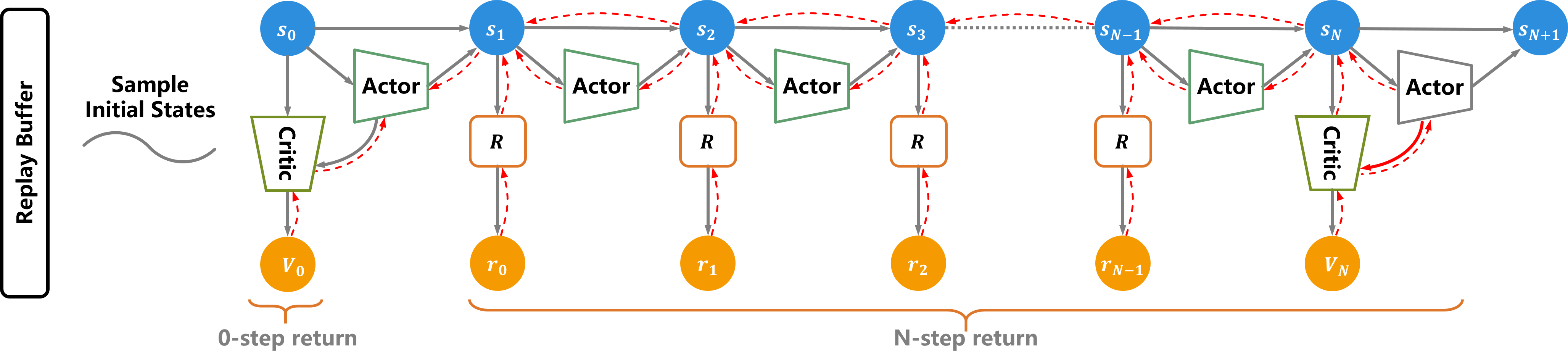}
    \caption{Overview of ABPT. ABPT combines 0-step return and N-step return to compensate for gradient bias resulting from partially non-differentiable rewards. The red dashed lines indicate the direction of backpropagation. The replay buffer stores only visited states for episode initialization to improve sampling efficiency and is not used for off-policy training.}
    \label{fig:ABPT}
\end{figure*}

\subsubsection{Sampling Insufficiency}
\label{sec:init}

\begin{figure}[t]
    \centering
    \includegraphics[width=0.8\linewidth]{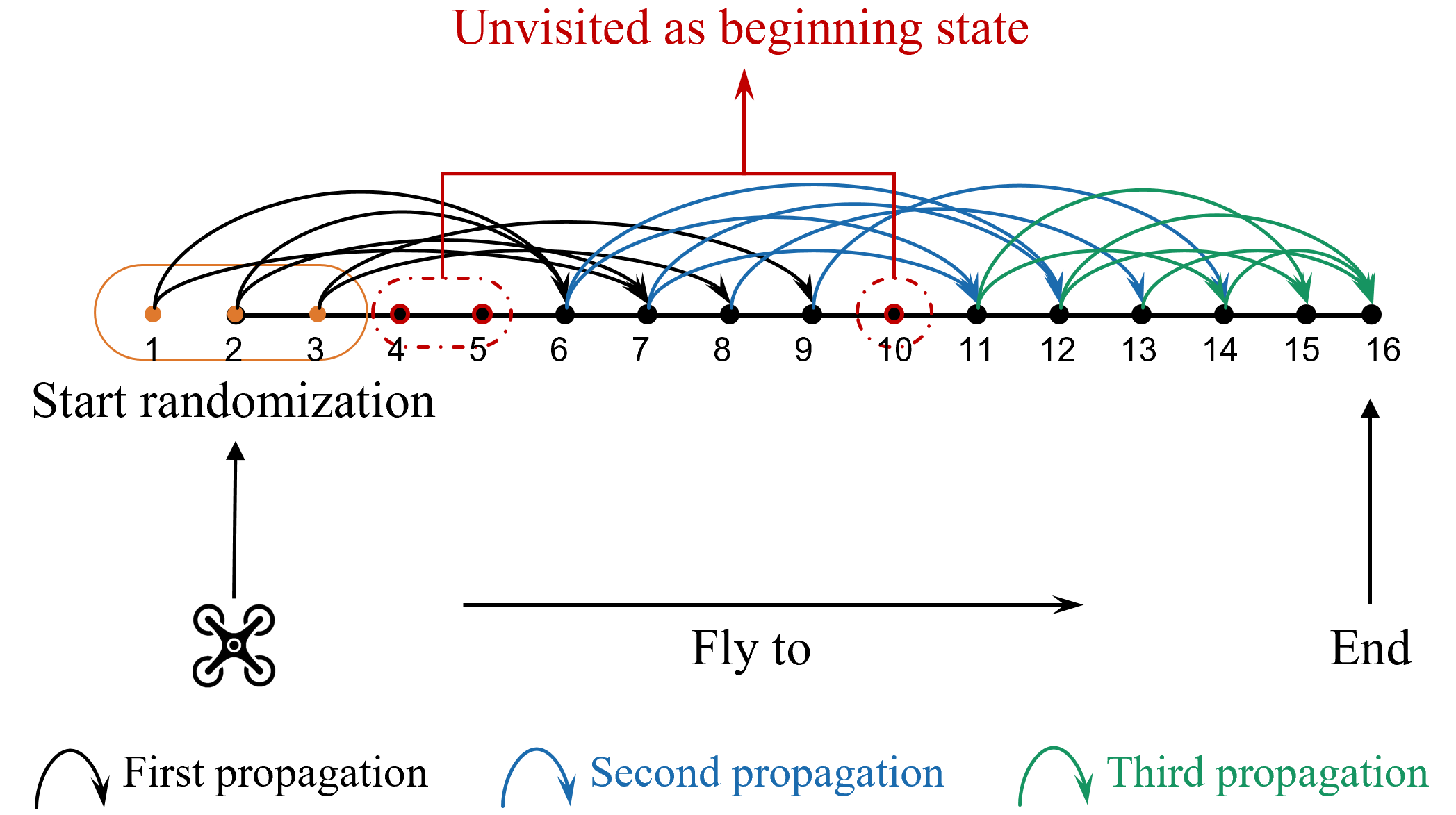}
    \caption{Illustration of the limited state coverage in the current BPTT implementation. Assuming the agent’s minimal horizon length is five, it cannot end at points 4 and 5 after executing a horizon even with randomization. As a result, part of the observation space never serves as beginning of horizons, reducing sampling efficiency.}
    \label{fig:state_init}
\end{figure}

For convenience, the current implementation of backpropagation-through-time (BPTT) in differentiable simulation always initializes the next computation horizon from the terminal state of the previous horizon. However, as \reffig{fig:state_init} shows, this design prevents certain states from ever being sampled as initial conditions , which leads to inefficient exploration of the observation space. In particular, states that are not reachable within a single horizon length cannot serve as starting points for training. This issue could be addressed by introducing an external replay buffer that records states at each step and resamples them as initial conditions, thereby improving coverage of the state space and enhancing sample efficiency. Noting that, in control task, the randomization domain could be enlarged enough to tackle such issue, but in planning task, it is usually constrained around the point of departure. Besides, regardless of randomization, the actual starting point distribution in observation space is still non-uniform, downgrading the training efficiency.

\subsubsection{Gradient Bias}

\begin{figure}[h]
    \centering
    \includegraphics[width=0.9\linewidth]{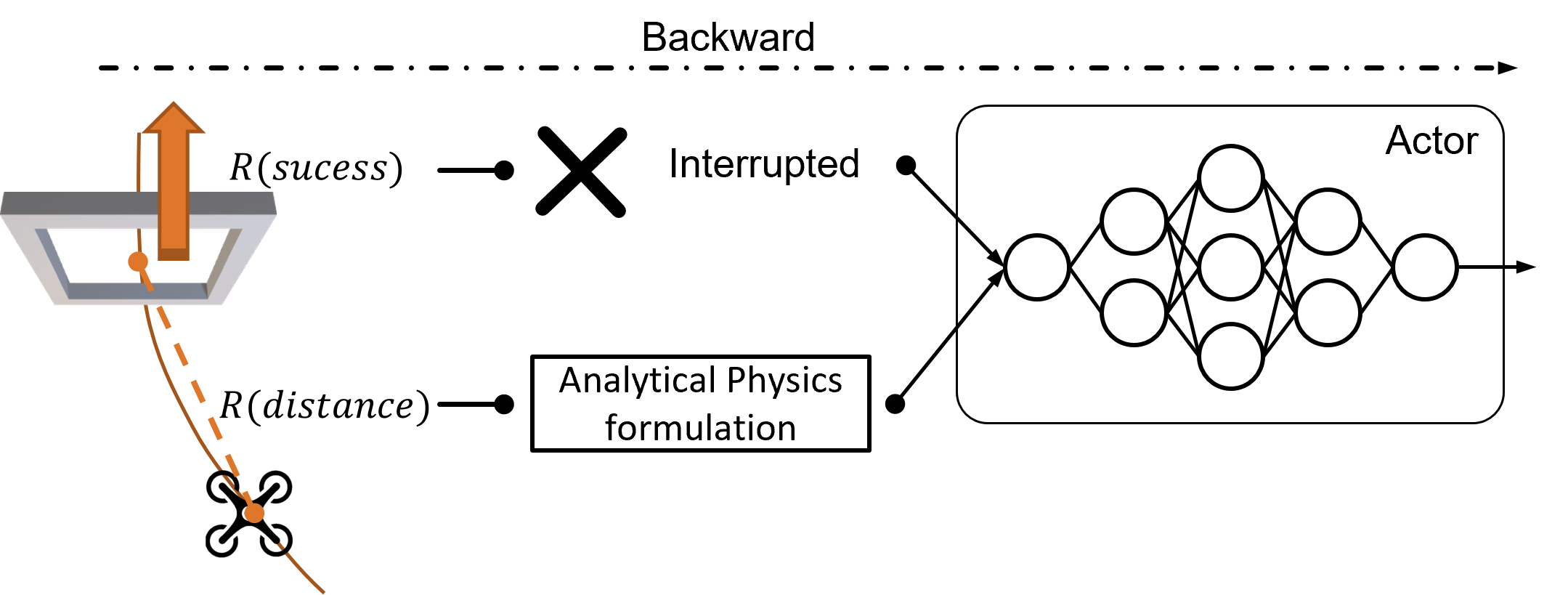}
    \caption{Illustration of gradient bias caused by partially non-differentiable rewards. In the racing task introduced in \refsec{sec:taskDefine}, the reward for passing a gate is a conditional constant and therefore does not directly contribute gradients through backpropagation.}
    \label{fig:biasGrad}
\end{figure}

When the rewards are partially differentiable, the gradients of non-differentiable part of the rewards will be absent from backpropagation.
For example, as shown in \reffig{fig:biasGrad}, a racing task's reward function consists of two components. The first one $R_{dist}$ depends on the distance from the drone to the gate to encourage the drone to move toward the gate, which is differentiable w.r.t the state. The second one $R_{succ}$ is a conditional constant score given for successfully passing the gate, which does not involve gradient computation w.r.t. policy parameters. Therefore, although the desired objective involves both rewards
\begin{equation}
\small
\mathcal{J}_\theta = \sum_{k=1}^{N} \gamma^k \Big(R_{dist}(s_{k})+\cancel{R_{succ}
(s_{k}})
\Big)
\end{equation}
backpropagation-through-time can effectively optimize only the differentiable components. 
As a result, the gate crossing reward $R_{succ}$, despite being crucial for learning the expected behavior (e.g. crossing the gate), is ignored during training. This ignorance can hinder the learned policy's ability to perform the desired actions. 
We refer to this phenomenon as gradient bias.


\subsection{Algorithm Formulation}

Motivated by the value gradient method, we propose to combine the 0-step return with N-step return for policy learning. This combination depresses the gradient bias while leveraging the strength of both gradient types. 
Our method, \textbf{A}mended \textbf{B}ackpropagation-through-\textbf{T}ime (ABPT), is an on-policy actor-critic learning approach.
An overview is presented in \reffig{fig:ABPT}.

During each training episode, we collect $|\mathcal{B}|$ trajectories with a horizon length $N$ and optimize the following objective function to update the actor network parameters $\theta$:

\begin{equation}
\label{eq:loss_actor}
\mathcal{J}_{\theta} = \frac{1}{2|\mathcal{B}|} \sum_{i=1}^{|\mathcal{B}|}  \Big(  \mathcal{J}_{\theta}^{N}+\mathcal{J}_{\theta}^{0} \Big)
\end{equation}
where $\mathcal{J}_{\theta}^{N}$, $\mathcal{J}_{\theta}^{0}$ are N-step return and 0-step return, defined as 
\begin{equation}
\small
\label{eq:n-return}
        \mathcal{J}_{\theta}^{N} = \left( \sum_{k = 0}^{N-1} \gamma^{k} {R}(s_{k}^i,a_{k}^i) \right) 
        + (1-d)\gamma^N Q_{\phi}(s_{N}^i,a_{N}^i) 
\end{equation}
\begin{equation}
\label{eq:0-return}
        \mathcal{J}_{\theta}^{0} =  Q_{\phi}
        (s_{0}^i,a_{0}^i)
\end{equation}

Here, $d$ is a boolean variable indicating whether the current episode has ended, and $i$ denotes the trajectory index.  Because each trajectory is generated by $\pi_\theta$, all terms are differentiable with respect to $\theta$. The first component in \refeq{eq:n-return} represents the accumulated reward within the horizon and the second is the prediction after the horizon, obtained by fixed critic. 
Both 0-step return and N-step return are expected values computed from the same action-value function $Q_\phi$.  
\refeq{eq:n-return} and \refeq{eq:0-return} respectively obtains the FOG and overall QG. Such combination not only precisely optimizes the policy via analytical physics, but also obtains the signal from binary rewards.  Here we empirically average two objects as \refeq{eq:loss_actor}, and the adaptive mixture will be studied in further work.


 

\begin{algorithm}[h]\
\caption{The proposed ABPT algorithm}
\label{alg:alg1}
\small
\begin{algorithmic}[1]
\STATE Initialize parameters $\phi, \phi^-, \theta$ randomly, initialize state buffer $\mathcal{D}=\{\}$.

\WHILE{num time-steps $<$ total time-steps}
    \STATE \# \textit{Train actor net}
    \STATE Sample minibatch $\{(s_i)\}_{\mathcal{B}} \sim \mathcal{D}$ as initial states
    \STATE Generate rollouts, save states in $\mathcal{D}$, compute the gradient of $\mathcal{J}_{\theta}$  and update the actor by gradient ascent $\theta\leftarrow\theta+ \alpha \nabla_\theta \mathcal{J}_{\theta} \hfill $ 
    \STATE
    \STATE \# \textit{Train critic net}
    \STATE Compute the estimated value $\tilde{Q}_\phi$ using \refeqn{eq:state_value_critic}
    \FOR {critic update step $c=1..C$}
        \STATE Compute the gradient of $\mathcal{L}_{\phi}$ and update weights by gradient descent $ \phi \leftarrow \phi - \alpha \nabla_\phi \mathcal{L}_{\phi}$ \hfill
        \STATE Softly update target critic $\phi^- \leftarrow (1 - \tau)\phi^- + \tau \phi$ \hfill 
    \ENDFOR

\ENDWHILE
\end{algorithmic}
\label{alg1}
\end{algorithm}
We use a Gaussian policy $\pi_\theta(a|s) = \mathcal{N}(\mu_\theta(s),\sigma_\theta(s))$ for the actor network and apply the reparameterization trick \cite{kingma2013auto} to gradient computation. We also normalize the actions using tanh function to stabilize the training process: $
    a_t = \tanh(\mu_{\theta}(s_t) + \sigma_{\theta}(s_t) \epsilon)$, where $ \epsilon \sim \mathcal{N}(0, I)$. After updating the critic, target returns are estimated over time and used to \zou{further refine the critic network parameters $\phi$} by minimizing the MSE loss function:
\begin{equation}
\small
\label{eq:loss_critic}
\mathcal{L}_{\phi} = \mathbb{E}_{s \in \{\tau_i\}} \left\| Q_{\phi}(s,a) - \tilde{Q}_{\phi}(s,a) \right\|^2.
\end{equation}
We employ TD$(\lambda)$ formulation \cite{sutton_reinforcement_2018} to estimate the expected return using exponentially averaging $k-$step returns:
\begin{equation}
\small
\label{eq:state_value_critic}
\tilde{Q}_{\phi}(s_t,a_t) = (1 - \lambda) \left( \sum_{k=1}^{N - t - 1} \lambda^{k - 1} G_t^k \right) + \lambda^{N - t - 1} G_t^{N - t}
\end{equation}
where $G_t^k$ denotes $k-$step return from $t$:
\begin{equation}
\small
\label{eq:return}
G_t^k = \left( \sum_{l=0}^{k-1} \gamma^l r_{t+l} \right) + (1-d)\gamma^k Q_{\phi}(s_{t+k}, a_{t+k}).
\end{equation}
where $d\in\{0,1\}$ indicates task termination.
To stabilize the critic training, we follow \cite{mnih2015human} to use a target critic $\phi^-$ to estimate the expected return (see \refeq{eq:state_value_critic}). 





 \zou{Existing methods~\cite{xu_accelerated_2022}} start each new horizon at the end of the previous horizon, which prevents certain regions of the state space from serving as initial states, resulting in inefficient sampling.
 To further encourage broader exploration during policy learning, we adopt a replay buffer to store all visited states throughout training. This buffer enables \textit{random sampling of dynamically feasible states for episode initialization}. While conceptually similar to the replay buffer used in off-policy learning algorithms, our approach differs in that we store only visited states rather than transitions, and use these states solely for initialization.


\section{Experiments}
\begin{figure*}[t]
    \centering
    \includegraphics[width=\textwidth]{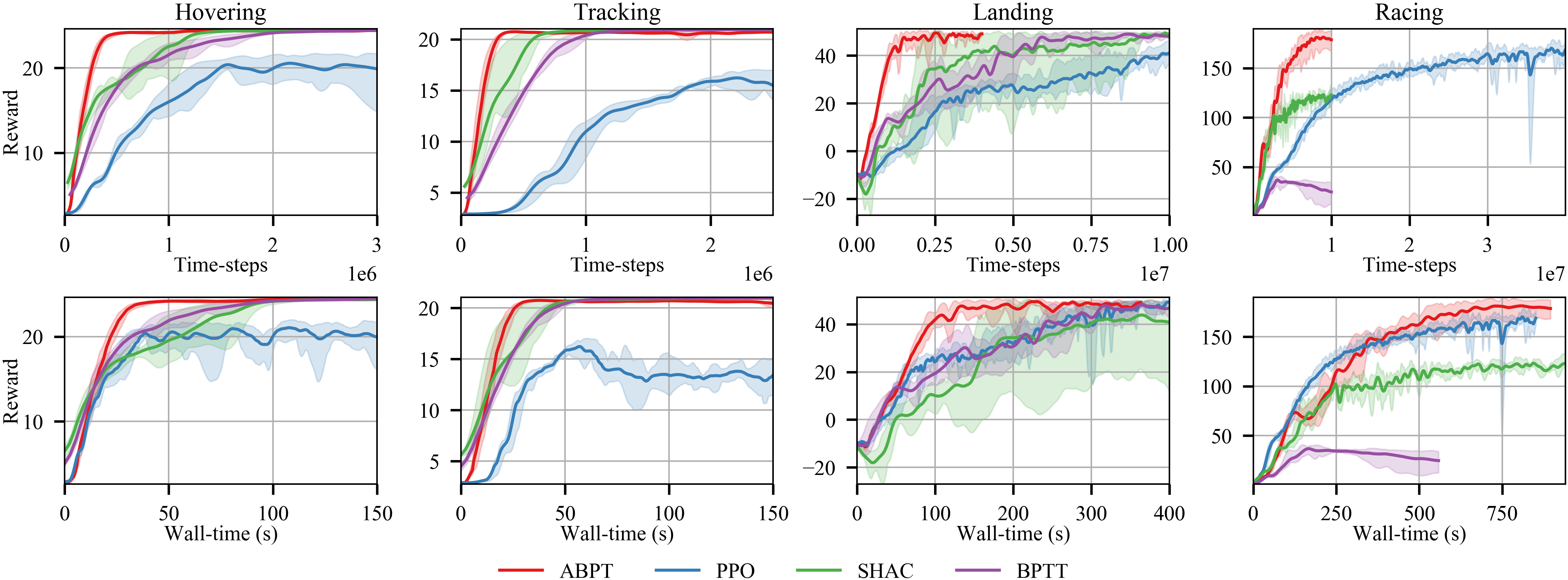}
    \caption{Training curves of PPO, SHAC, BPTT, and our ABPT in both time-step (\textbf{Top}) and wall-time (\textbf{Bottom}). Each curve is averaged over results from five random seeds, and the shaded area denotes the range of best and worst reward.}
    \label{fig:std_compa}
    \vspace{-10pt}
\end{figure*}

\subsection{Experiment Setup}
\label{sec:taskDefine}



In our experiments, we evaluate the proposed ABPT against three widely used baseline methods: PPO~\cite{schulman2017proximal}, BPTT~\cite{freeman_brax_2021}, and SHAC~\cite{xu_accelerated_2022}, which are all the representatives of its category.

To ensure fair comparison, we implemented SHAC and BPTT by ourselves based on available source code, and adopt PPO from stable-baselines3 \cite{raffin2021stable} in VisFly simulator. SHAC and BPTT are implemented strictly upon stable-baselines3 standard for following convenient development. All algorithms used parallel differentiable simulations to accelerate training. We tuned all hyperparameters to achieve optimal performance, and kept the settings consistent across all experiments as much as possible. 
All experiments were conducted with 5 random seeds for validation of robustness. Given the different time-step metrics across the algorithms, we compare their performance in terms of wall-time as well. \reffig{fig:std_compa} provides reward curves of all methods during training.


\subsection{Training Results on Multi-Task Quadrotor Control}

\underline{PPO}: 
PPO demonstrates moderate performance across the four tasks. However, due to the lack of an analytical gradient, PPO requires more sample collections to estimate the policy gradient, making it slower in terms of time-steps. In tasks that involve fully differentiable rewards such as hovering and tracking, it achieves the lowest ultimate reward compared to FOG-based algorithms. As expected, PPO produces smooth and acceptable learning curves, since non-differentiable rewards do not impact the ZOG used by PPO.

\underline{BPTT}: BPTT exhibits similar performance to SHAC and ABPT in the first two tasks. In the Landing task, despite the reward function incorporating non-differentiable discrete scores upon success, this component has only a minor impact on the FOG computation. This is because the reward function excluding this constant, has correctly determined the gradient via backpropagation. In the Racing task, we apply learning rate decay to BPTT, SHAC, and ABPT. BPTT shows the worst performance among all algorithms, demonstrating that the iteration quickly converges to a local minimum, caused by the bias introduced by the non-differentiable part in rewards. 

\underline{SHAC}: Even though FOG is minimally biased in the Landing task, the curves from the five random seeds show significant fluctuations. The terminal success reward leads to high variance in the TD$(\lambda)$ formulation used to estimate N-step returns, complicating critic training. As a result, SHAC performs worse than BPTT in the Landing task. In the Racing task, the terminal value partially addresses the non-differentiable components but still performs much worse than PPO and ABPT.

\underline{ABPT}: In all tests, ABPT method converges to the highest rewards. It achieves the fastest convergence speeds in the first three tasks and similar convergence speed to PPO in the racing tasks. 
By replaying visited states as initial states, ABPT enhances sampling efficiency by broadening coverage over previously visited but under-sampled feasible states. 
In the racing task, ABPT also outperforms PPO with a higher converged reward. This is largely because the value gradient introduced by the 0-step return is not directly blocked by the non-differentiable reward term, making ABPT effective at compensating for gradient bias.



\subsection{Ablation}
\label{sec:ablation}


\begin{figure}[h]
  \vspace{-12pt} 
  \centering
  \includegraphics[width=\linewidth]{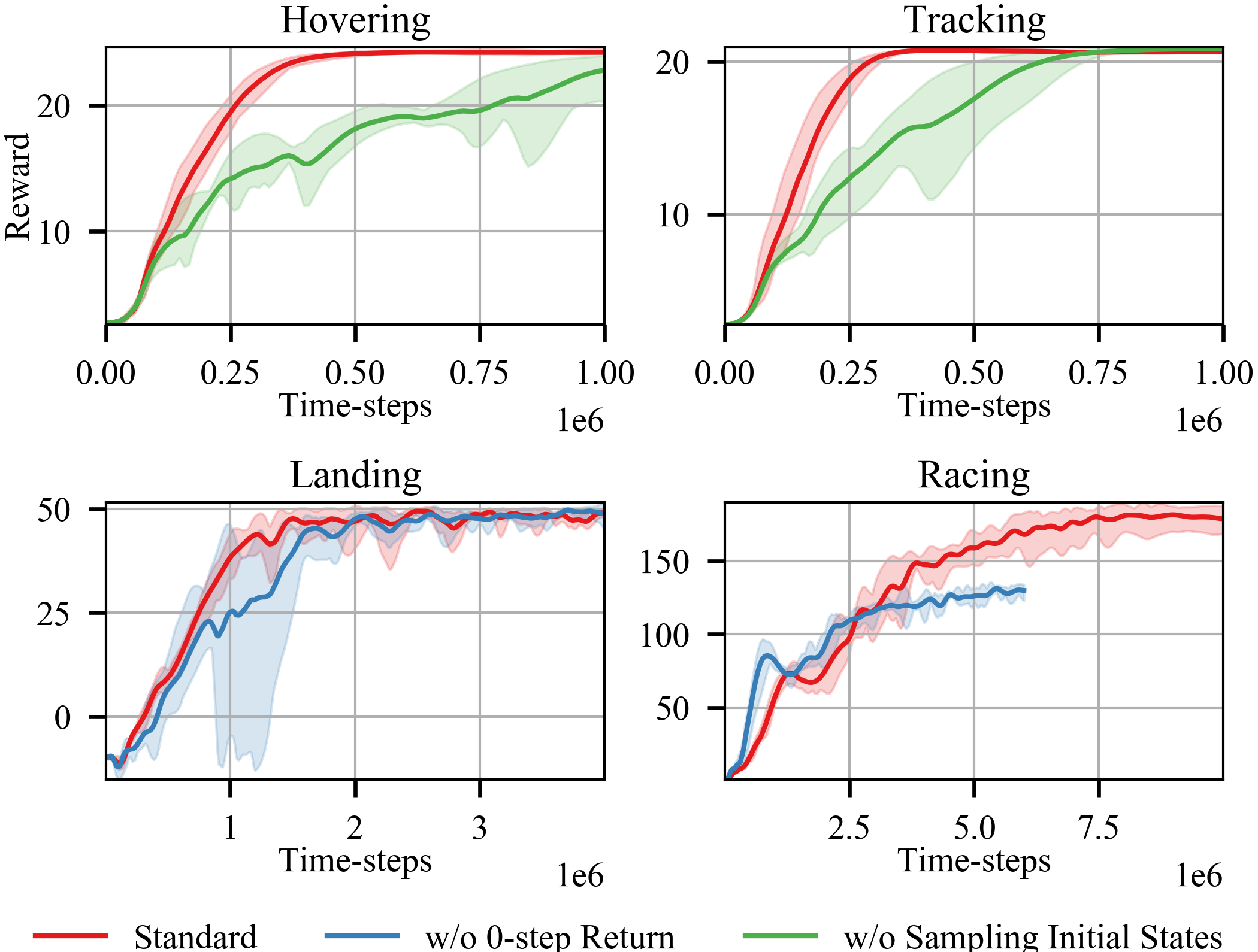}
  \caption{Ablation study: the key components of ABPT are sequentially removed in turn to evaluate each one's contribution.}
  \label{fig:ablation}
  \vspace{-12pt} 
\end{figure}

As shown in \reffig{fig:ablation}, we evaluate the effectiveness of key components of our approach by removing each during training. The results show that:
1) Incorporating 0-step return clearly improves the final converged performance in tasks with non-differentiable rewards such as landing and racing. 

2) Initializing episodes from previously visited states stored in the buffer enhances sampling efficiency, accelerating convergence. 
3) Removing the N-step return significantly reduce landing performance but has negligible impact on racing. In racing, the binary reward primarily drives the gradient, whereas in landing it serves only as an auxiliary guidance. This suggests that N-step return does not substantially contribute to mitigating biased gradients.

    

\subsection{Discontinuity Relaxation}
\begin{figure}[h]
    \centering
    \includegraphics[width=0.9\linewidth]{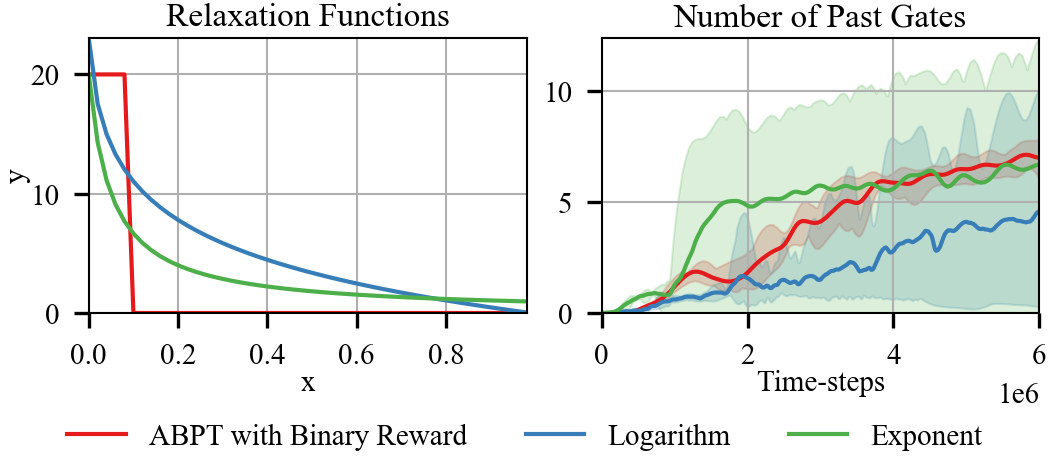}
    \caption{Training curves for the racing task using binary reward relaxations, which are trained with SHAC after replacing the binary reward with relaxed continuous reward.}
    \label{fig:relax}
\end{figure}

To evaluate the effectiveness of relaxation techniques in addressing non-differentiable rewards, we replace the binary reward with smooth approximations that closely resemble its behavior \zou{while using common BPTT}. Specifically, we employ logarithmic ($-5\log(|p-\hat{p}|+0.01)$) and exponential ($1/(|p-\hat{p}|+0.05)$) relaxations, as shown in \reffig{fig:relax}. Both functions exhibit a similar trend to the original binary reward.

The objective of racing task is to pass through as many gates as possible. Since the reward scales of different relaxations are not directly comparable, the number of passed gates provides a fairer metric. Although the exponential relaxation achieves performance comparable to ABPT, its variance is significantly higher, leading to training instability. As a result, the quadrotor is more likely to become trapped in local minima. Such relaxation encourages the quadrotor to hover near a gate to repeatedly obtain sub-optimal rewards rather than flying forward to the next gate.

\subsection{Reward Robustness}
\label{sec:reward_robustness}

\begin{figure}[h]
    \centering
    \includegraphics[width=0.9\linewidth]{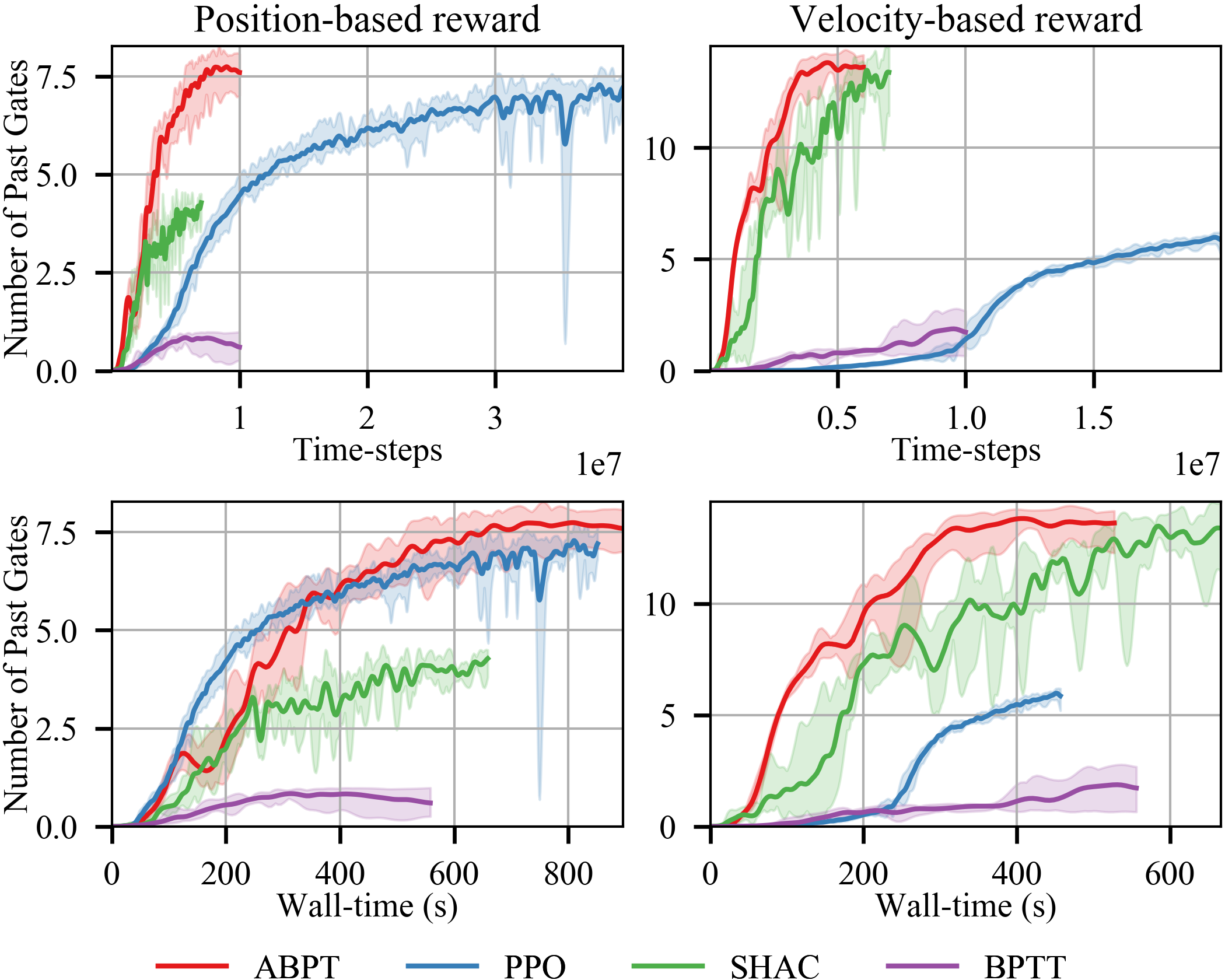}
    \caption{Training curves with different rewards: position-based rewards (\textbf{Left column}) and velocity-based rewards (\textbf{Right column}). The number of passed gates is visualized as the performance metric because of different rewards used for training.
    }
    \label{fig:rewardrobust}
\end{figure}

Designing an appropriate reward function is highly challenging for real-world applications, particularly when dealing with specific requirements. Ensuring robustness to reward architecture is crucial for the training algorithms. In the racing task, we redefined the reward function by replacing Euclidean distance with approaching velocity in the reward. 
As shown in \reffig{fig:rewardrobust}, ABPT outperforms other methods with both position-based and velocity-based rewards. With fewer non-differentiable components, velocity-based rewards allow ABPT and SHAC to pass more gates per episode, while BPTT fails due to gradient issues.

\subsection{Proof-of-Concept Real-World Deployment}
\label{sec:realWorldExp}
We further provide proof-of-concept real-world deployments of the learned policies. As illustrated in Fig.~1 and the supplementary video, policies trained in the proposed framework were deployed on real quadrotor platforms and were able to complete four representative tasks, namely hovering, tracking, landing, and racing. These deployments qualitatively demonstrate that the proposed framework captures practically relevant task structure and dynamics to support transfer beyond simulation. We emphasize that this experiment is intended to demonstrate initial transferability rather than superiority over alternative deployment pipelines.


\section{Discussion and Limitation}
\label{sec:discuss}
The results indicate that the proposed framework provides a unified testbed for first-order reinforcement learning across multiple quadrotor tasks, while ABPT improves robustness particularly in partially differentiable settings. The hardware results are intended as proof-of-concept evidence of deployability and initial sim-to-real transferability, rather than a comprehensive quantitative benchmark. In future work, we will conduct more comprehensive quantitative evaluations and further investigate the method in greater detail. The source code is released at \url{https://anonymous.4open.science/r/APG-E73E}.

\section{Conclusion}
This paper presented a unified differentiable framework and ABPT for first-order reinforcement learning of quadrotor control. The proposed framework provides a wrapped and extensible setting for developing and evaluating first-order RL methods across representative quadrotor tasks, including hovering, tracking, landing, and racing. Within this setting, we identified two practical bottlenecks of standard first-order training: limited state coverage caused by horizon initialization, and gradient bias caused by partially non-differentiable rewards. To address these issues, we proposed Amended Backpropagation Through Time (ABPT), which combines short-horizon differentiable optimization, a value-based auxiliary objective, and visited-state initialization. Experimental results showed that ABPT is competitive across the unified benchmark, with the clearest gains in partially non-differentiable tasks. We further provided proof-of-concept real-world deployments, offering initial evidence that policies learned in the proposed framework can transfer beyond simulation.



\bibliography{reference}  

@book{sutton_reinforcement_2018,
	title = {Reinforcement {Learning}, second edition: {An} {Introduction}},
	isbn = {978-0-262-35270-3},
	shorttitle = {Reinforcement {Learning}, second edition},
	language = {en},
	publisher = {MIT Press},
	author = {Sutton, Richard S. and Barto, Andrew G.},
	month = nov,
	year = {2018},
	note = {Google-Books-ID: uWV0DwAAQBAJ},
}

@article{mnih2015human,
  title={Human-level control through deep reinforcement learning},
  author={Mnih, Volodymyr and Kavukcuoglu, Koray and Silver, David and Rusu, Andrei A and Veness, Joel and Bellemare, Marc G and Graves, Alex and Riedmiller, Martin and Fidjeland, Andreas K and Ostrovski, Georg and others},
  journal={nature},
  volume={518},
  number={7540},
  pages={529--533},
  year={2015},
  publisher={Nature Publishing Group UK London}
}

@misc{xu_accelerated_2022,
	title = {Accelerated {Policy} {Learning} with {Parallel} {Differentiable} {Simulation}},
	
	doi = {10.48550/arXiv.2204.07137},
	urldate = {2024-10-12},
	publisher = {arXiv},
	author = {Xu, Jie and Makoviychuk, Viktor and Narang, Yashraj and Ramos, Fabio and Matusik, Wojciech and Garg, Animesh and Macklin, Miles},
	month = apr,
	year = {2022},
	note = {arXiv:2204.07137},
}

@article{kingma2013auto,
  title={Auto-encoding variational bayes},
  author={Kingma, Diederik P},
  publisher={arXiv preprint arXiv:1312.6114},
  year={2013}
}

@inproceedings{gao_adaptive_gradient_2024,
	title = {Adaptive-{Gradient} {Policy} {Optimization}: {Enhancing} {Policy} {Learning} in {Non}-{Smooth} {Differentiable} {Simulations}},
	shorttitle = {Adaptive-{Gradient} {Policy} {Optimization}},
	
	language = {en},
	urldate = {2024-11-16},
	booktitle = {Proceedings of the 41st {International} {Conference} on {Machine} {Learning}},
	publisher = {PMLR},
	author = {Gao, Feng and Shi, Liangzhi and Zhang, Shenao and Wang, Zhaoran and Wu, Yi},
	month = jul,
	year = {2024},
	note = {ISSN: 2640-3498},
	pages = {14844--14858},
}

@inproceedings{mora_pods_2021,
	title = {{PODS}: {Policy} {Optimization} via {Differentiable} {Simulation}},
	shorttitle = {{PODS}},
	
	language = {en},
	urldate = {2024-11-17},
	booktitle = {Proceedings of Vthe 38th {International} {Conference} on {Machine} {Learning}},
	publisher = {PMLR},
	author = {Mora, Miguel Angel Zamora and Peychev, Momchil and Ha, Sehoon and Vechev, Martin and Coros, Stelian},
	month = jul,
	year = {2021},
	note = {ISSN: 2640-3498},
	pages = {7805--7817},
}

@article{zhang2024back,
  title={Back to Newton's Laws: Learning Vision-based Agile Flight via Differentiable Physics},
  author={Zhang, Yuang and Hu, Yu and Song, Yunlong and Zou, Danping and Lin, Weiyao},
  journal={arXiv preprint arXiv:2407.10648},
  year={2024}
}

@inproceedings{n_wiedemann_training_2023,
	title = {Training {Efficient} {Controllers} via {Analytic} {Policy} {Gradient}},
	doi = {10.1109/ICRA48891.2023.10160581},
	author = {N. Wiedemann and V. Wüest and A. Loquercio and M. Müller and D. Floreano and D. Scaramuzza},
	month = jun,
	year = {2023},
	pages = {1349--1356},
}

@article{schulman2017proximal,
  title={Proximal policy optimization algorithms},
  author={Schulman, John and Wolski, Filip and Dhariwal, Prafulla and Radford, Alec and Klimov, Oleg},
  journal={arXiv preprint arXiv:1707.06347},
  year={2017}
}

@inproceedings{heiden_neuralsim_2021,
	title = {{NeuralSim}: {Augmenting} {Differentiable} {Simulators} with {Neural} {Networks}},
	shorttitle = {{NeuralSim}},
	
	doi = {10.1109/ICRA48506.2021.9560935},
	urldate = {2024-11-27},
	booktitle = {2021 {IEEE} {International} {Conference} on {Robotics} and {Automation} ({ICRA})},
	author = {Heiden, Eric and Millard, David and Coumans, Erwin and Sheng, Yizhou and Sukhatme, Gaurav S.},
	month = may,
	year = {2021},
	note = {ISSN: 2577-087X},
	pages = {9474--9481},
}

@misc{li_visfly_2024,
	title = {{VisFly}: {An} {Efficient} and {Versatile} {Simulator} for {Training} {Vision}-based {Flight}},
	shorttitle = {{VisFly}},
	
	doi = {10.48550/arXiv.2407.14783},

	urldate = {2024-11-27},
	publisher = {arXiv},
	author = {Li, Fanxing and Sun, Fangyu and Zhang, Tianbao and Zou, Danping},
	month = sep,
	year = {2024},
	note = {arXiv:2407.14783},
}

@inproceedings{schoenholz_jax_2020,
	title = {{JAX} {MD}: {A} {Framework} for {Differentiable} {Physics}},
	volume = {33},
	shorttitle = {{JAX} {MD}},
	urldate = {2024-11-27},
	booktitle = {Advances in {Neural} {Information} {Processing} {Systems}},
	publisher = {Curran Associates, Inc.},
	author = {Schoenholz, Samuel and Cubuk, Ekin Dogus},
	year = {2020},
	pages = {11428--11441},
}

@misc{freeman_brax_2021,
	title = {Brax -- {A} {Differentiable} {Physics} {Engine} for {Large} {Scale} {Rigid} {Body} {Simulation}},
	
	doi = {10.48550/arXiv.2106.13281},
	urldate = {2024-11-27},
	publisher = {arXiv},
	author = {Freeman, C. Daniel and Frey, Erik and Raichuk, Anton and Girgin, Sertan and Mordatch, Igor and Bachem, Olivier},
	month = jun,
	year = {2021},
	note = {arXiv:2106.13281},
}

@misc{hu_difftaichi_2020,
	title = {{DiffTaichi}: {Differentiable} {Programming} for {Physical} {Simulation}},
	shorttitle = {{DiffTaichi}},
	
	doi = {10.48550/arXiv.1910.00935},
	urldate = {2024-11-27},
	publisher = {arXiv},
	author = {Hu, Yuanming and Anderson, Luke and Li, Tzu-Mao and Sun, Qi and Carr, Nathan and Ragan-Kelley, Jonathan and Durand, Frédo},
	month = feb,
	year = {2020},
	note = {arXiv:1910.00935},
}

@misc{howell_dojo_2023,
	title = {Dojo: {A} {Differentiable} {Physics} {Engine} for {Robotics}},
	shorttitle = {Dojo},
	
	doi = {10.48550/arXiv.2203.00806},
	urldate = {2024-11-27},
	publisher = {arXiv},
	author = {Howell, Taylor A. and Cleac'h, Simon Le and Brüdigam, Jan and Kolter, J. Zico and Schwager, Mac and Manchester, Zachary},
	month = mar,
	year = {2023},
	note = {arXiv:2203.00806},
}

@inproceedings{todorov_mujoco_2012,
	title = {{MuJoCo}: {A} physics engine for model-based control},
	shorttitle = {{MuJoCo}},
	
	doi = {10.1109/IROS.2012.6386109},
	urldate = {2024-11-27},
	booktitle = {2012 {IEEE}/{RSJ} {International} {Conference} on {Intelligent} {Robots} and {Systems}},
	author = {Todorov, Emanuel and Erez, Tom and Tassa, Yuval},
	month = oct,
	year = {2012},
	note = {ISSN: 2153-0866},
	pages = {5026--5033},
}

@article{paszke2017automatic,
  title={Automatic differentiation in pytorch},
  author={Paszke, Adam and Gross, Sam and Chintala, Soumith and Chanan, Gregory and Yang, Edward and DeVito, Zachary and Lin, Zeming and Desmaison, Alban and Antiga, Luca and Lerer, Adam},
  year={2017}
}

@inproceedings{savva2019habitat,
  title={Habitat: A platform for embodied ai research},
  author={Savva, Manolis and Kadian, Abhishek and Maksymets, Oleksandr and Zhao, Yili and Wijmans, Erik and Jain, Bhavana and Straub, Julian and Liu, Jia and Koltun, Vladlen and Malik, Jitendra and others},
  booktitle={Proceedings of the IEEE/CVF international conference on computer vision},
  pages={9339--9347},
  year={2019}
}

@article{raffin2021stable,
  title={Stable-baselines3: Reliable reinforcement learning implementations},
  author={Raffin, Antonin and Hill, Ashley and Gleave, Adam and Kanervisto, Anssi and Ernestus, Maximilian and Dormann, Noah},
  journal={Journal of Machine Learning Research},
  volume={22},
  number={268},
  pages={1--8},
  year={2021}
}

@misc{georgiev2024adaptivehorizonactorcriticpolicy,
      title={Adaptive Horizon Actor-Critic for Policy Learning in Contact-Rich Differentiable Simulation}, 
      author={Ignat Georgiev and Krishnan Srinivasan and Jie Xu and Eric Heiden and Animesh Garg},
      year={2024},
      eprint={2405.17784},
      archivePrefix={arXiv},
      primaryClass={cs.LG},
       
}

@article{xing2024stabilizing,
  title={Stabilizing reinforcement learning in differentiable multiphysics simulation},
  author={Xing, Eliot and Luk, Vernon and Oh, Jean},
  journal={arXiv preprint arXiv:2412.12089},
  year={2024}
}

@inproceedings{heeg2025learning,
  title={Learning quadrotor control from visual features using differentiable simulation},
  author={Heeg, Johannes and Song, Yunlong and Scaramuzza, Davide},
  booktitle={2025 IEEE International Conference on Robotics and Automation (ICRA)},
  pages={4033--4039},
  year={2025},
  organization={IEEE}
}

@misc{lu_yopov2-tracker_2025,
	title = {{YOPOv}2-Tracker: An End-to-End Agile Tracking and Navigation Framework from Perception to Action},
	
	doi = {10.48550/arXiv.2505.06923},
	shorttitle = {{YOPOv}2-Tracker},
	number = {{arXiv}:2505.06923},
	publisher = {{arXiv}},
	author = {Lu, Junjie and Hui, Yulin and Zhang, Xuewei and Feng, Wencan and Shen, Hongming and Li, Zhiyu and Tian, Bailing},
	urldate = {2025-07-09},
	date = {2025-05-11},
	eprinttype = {arxiv},
	eprint = {2505.06923 [cs]},
}

@misc{pan_learning_2025,
	title = {Learning on the {Fly}: {Rapid} {Policy} {Adaptation} via {Differentiable} {Simulation}},
	shorttitle = {Learning on the {Fly}},
	
	doi = {10.48550/arXiv.2508.21065},
	urldate = {2025-09-01},
	publisher = {arXiv},
	author = {Pan, Jiahe and Xing, Jiaxu and Reiter, Rudolf and Zhai, Yifan and Aljalbout, Elie and Scaramuzza, Davide},
	month = aug,
	year = {2025},
	note = {arXiv:2508.21065 [cs]},
}
\bibliographystyle{IEEEtran}

\clearpage

\end{document}